\documentclass[11pt]{article}

\usepackage[final]{acl}

\usepackage{times}
\usepackage{latexsym}
\usepackage{amsmath}
\usepackage[T1]{fontenc}
\usepackage{multirow}
\usepackage{booktabs} % \documentclass와 \begin{document} 사이에 추가해야 합니다.
\usepackage{amsthm}
\usepackage{xcolor}
 
\usepackage[utf8]{inputenc}

\usepackage{microtype}
\usepackage{amssymb}
\usepackage{inconsolata}

\usepackage{graphicx}
\usepackage{xspace}
\usepackage{orcidlink}
\newcommand{\eg}{\textit{e.g.,}\xspace}
\newcommand{\ie}{\textit{i.e.,}\xspace}

\title{CS-CLIP: Compositional Scene Graph-guided CLIP for Robust Compositional Reasoning}
\author{SeongJun Jeong\,\orcidlink{0009-0007-6672-2419} \quad
  Minjoon Jung\,\orcidlink{0000-0003-0838-7687} \quad
  Woo Suk Choi\,\orcidlink{0009-0001-8091-347X} \\
  {\bf Youwon Jang\,\orcidlink{0000-0001-6714-3717} \quad
  Byoung-Tak Zhang\,\orcidlink{0000-0001-9890-0389}\thanks{\;Corresponding author.}} \\
  Seoul National University \\
  \texttt{\{sjjeong, mjjung, wschoi, ywjang, btzhang\}@bi.snu.ac.kr}}
\begin{document}
\maketitle

\begin{abstract}
Vision-language models (VLMs) demonstrate strong performance across compositional reasoning benchmarks, which require reasoning over semantic perturbations of objects, attributes, relations, and their interactions.
However, our controlled analysis reveals that existing compositionality-aware VLMs exhibit element-specific biases, often underperforming vanilla CLIP on certain compositional elements.
To address this, we propose Compositional Scene Graph-guided CLIP (CS-CLIP), which uses scene graphs to identify compositional elements and construct structured negatives via selective masking.
We further retain negatives that are most contradictory to the original caption, forcing the model to rely on compositional structure rather than surface cues.
CS-CLIP achieves state-of-the-art compositional reasoning with robust performance across compositional elements. 
It also preserves general vision-language capabilities such as cross-modal retrieval and downstream visual reasoning, while requiring fewer training samples than prior methods.
\end{abstract}

\section{Introduction}

Vision-language models (VLMs) (\eg CLIP~\cite{radford2021learning}) learn transferable multimodal representations that serve as effective foundations.
With this advancement, compositionality-aware VLMs~\cite{yuksekgonul2022and, huang2024structure, patel2024tripletclip} have emerged, enabling models to reason over compositional elements and their interactions by incorporating hard negative captions into contrastive training to improve compositional reasoning.

\begin{figure}[t]
    \centering
    % 검은색 박스로 공간 차지 (너비: 컬럼너비, 높이: 7cm)
    % \framebox[\linewidth]{\rule{0pt}{7cm}}
    \includegraphics[width=\linewidth]{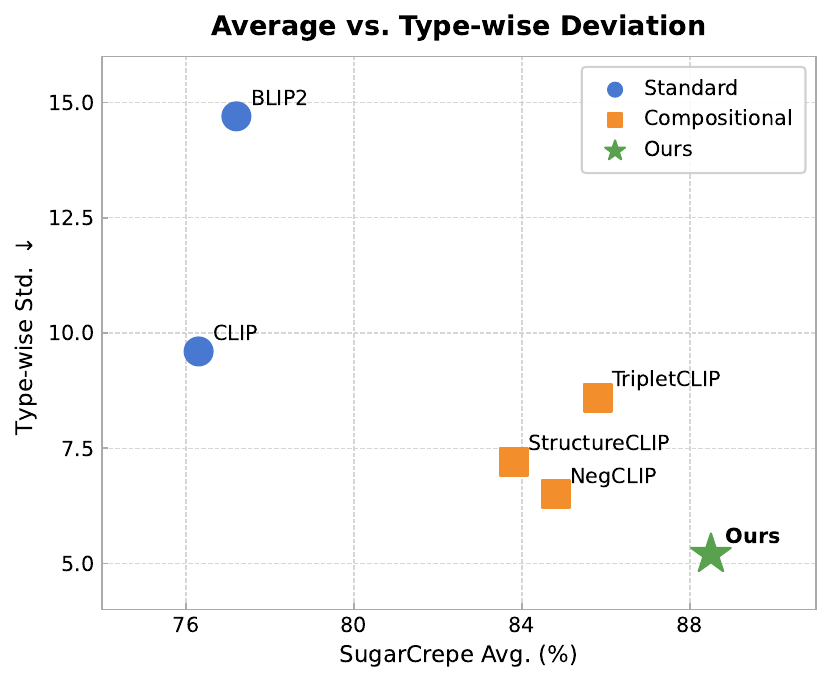}
    \caption{\textbf{Average performance versus type-wise deviation on SugarCrepe.}
    CS-CLIP (Ours) achieves the highest SugarCrepe average with the lowest standard deviation across reasoning types, indicating more robust compositional reasoning.}
    \label{fig:teaser}
\end{figure}

Despite their reported gains on compositional benchmarks~\cite{yuksekgonul2022and, hsieh2023sugarcrepe, dumpala2024sugarcrepe++}, it remains unclear how well such methods handle individual compositional elements.
Evaluating element-specific behavior requires test cases that isolate the targeted compositional element from other cues.
We construct such controlled test cases by fixing the perturbation template and substitute vocabulary, isolating a single compositional element per instance.
We measure both \textit{accuracy} and \textit{sensitivity} to evaluate whether models can confidently capture changes in compositional elements. 
Our results reveal that existing compositionality-aware VLMs fail to demonstrate clear improvements over the vanilla CLIP backbone; in fact, they frequently underperform it across different element changes.
This indicates a significant deficiency and bias in existing VLMs, which have not surfaced in previous benchmarks. 

To address this, we propose Compositional Scene Graph-guided CLIP (CS-CLIP), which generates hard negatives that explicitly target individual compositional elements. 
CS-CLIP parses captions into scene graphs~\cite{johnson2015image, wu2019unified} to localize objects, attributes, and relations, and applies a diverse set of perturbations—replacement, addition, and swapping—to each element independently, ensuring that supervision covers every element type. 
To produce negatives that are distinguishable from the positive only through compositional reasoning, we adopt a generate-and-rank strategy: a masked language model (MLM)~\cite{devlin2019bert} generates fluent candidates, and a natural language inference model (NLI)~\cite{liu2019roberta} selects those that semantically contradict the original caption. 
Training on these element-targeted hard negatives directly addresses the biases identified above, encouraging the model to reason over compositional structure rather than rely on superficial cues.

On compositional reasoning benchmarks~\cite{hsieh2023sugarcrepe, yuksekgonul2022and}, CS-CLIP achieves state-of-the-art average performance with the lowest type-wise deviation, indicating robust compositional reasoning (Figure~\ref{fig:teaser}).
CS-CLIP also preserves competitive cross-modal retrieval on COCO~\cite{lin2014microsoft}, transfers to zero-shot visual reasoning on GQA~\cite{hudson2019gqa}, and achieves its compositional gains with substantially fewer training samples than prior methods.

Our contributions are threefold:

\begin{itemize}
    \item We reveal element-specific biases in existing compositionality-aware VLMs, which even underperform vanilla CLIP on certain elements, through our controlled test cases.
    \item We propose CS-CLIP, which constructs element-targeted hard negatives via scene graph parsing and a generate-and-rank pipeline ensuring fluency and semantic contradiction.
    \item CS-CLIP achieves state-of-the-art compositional reasoning, mitigating element-specific biases while preserving general vision-language capabilities with substantially fewer training samples.
\end{itemize}

\begin{figure*}[th!]
    \centering
    % \framebox[\textwidth]{\rule{0pt}{8cm}}
    \includegraphics[width=\textwidth]{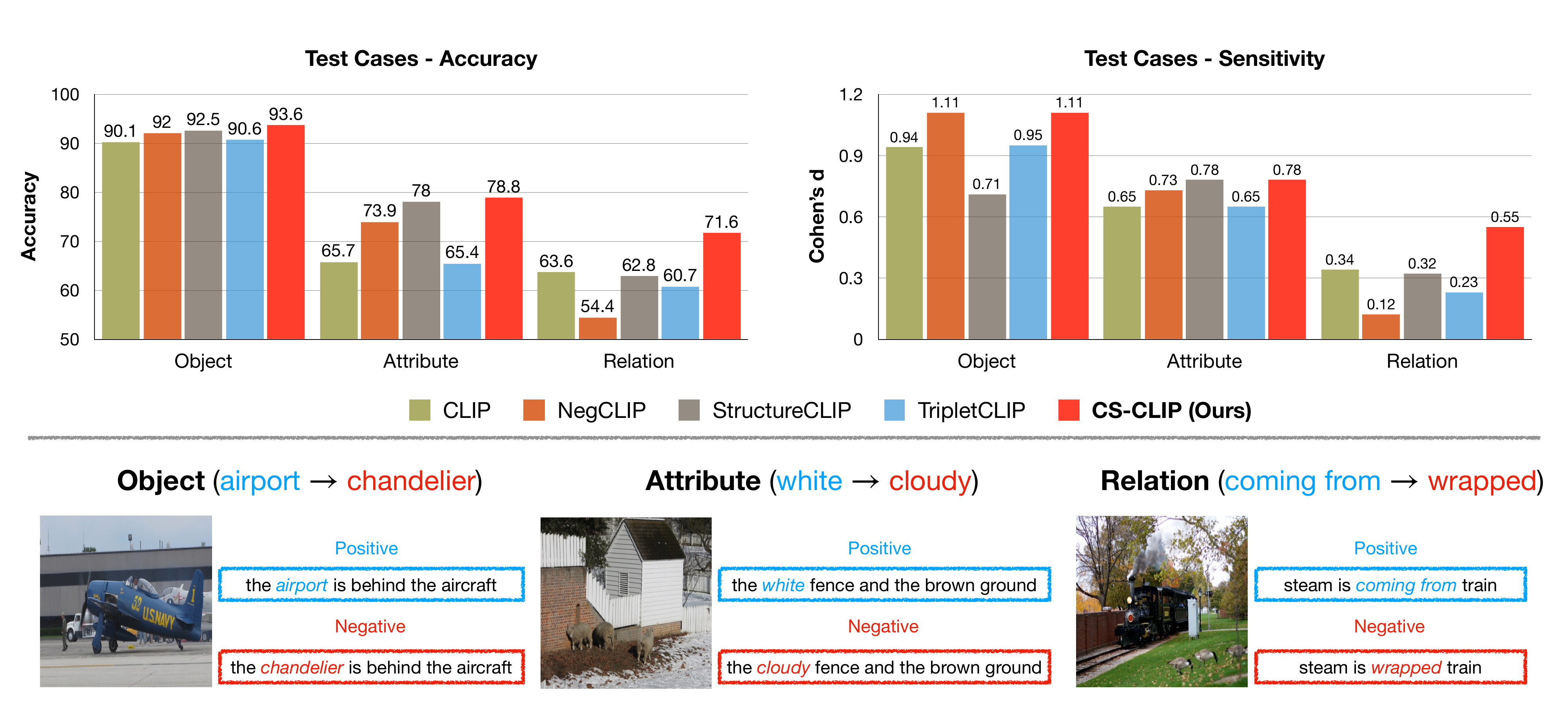}
    \caption{\textbf{Results of our controlled evaluation.} We illustrate the accuracy and sensitivity metrics for each element (top) and present representative cases of positive and hard negative captions (bottom).}
    \label{fig:core_analysis}
\end{figure*}

\section{Related Work}

\noindent \textbf{Vision-Language Models.}
Vision-language models (VLMs) trained via large-scale contrastive learning~\cite{radford2021learning, jia2021scaling, li2021align, li2022blip, li2023blip} align image and text representations into a shared embedding space, demonstrating strong generalization across various downstream tasks~\cite{lin2014microsoft, plummer2015flickr30k, deng2009imagenet, song2022clip, rombach2022high, zhou2023zegclip}.
However, these models often exhibit bag-of-words behavior~\cite{yuksekgonul2022and, thrush2022winoground}, prioritizing salient keywords over fine-grained compositional structure~\cite{zhao2022vl, hsieh2023sugarcrepe}, which limits their compositional reasoning across objects, attributes, and relations.

\noindent \textbf{Compositional Reasoning Benchmarks.}
Several benchmarks have been proposed to assess compositional reasoning in VLMs~\cite{yuksekgonul2022and, thrush2022winoground, zhao2022vl, hsieh2023sugarcrepe, dumpala2024sugarcrepe++}.
Early benchmarks such as VL-Checklist~\cite{zhao2022vl} and ARO~\cite{yuksekgonul2022and} construct hard negatives via rule-based perturbations, which can introduce artifacts that language-only models exploit through text-only plausibility cues~\cite{hsieh2023sugarcrepe}.
SugarCrepe~\cite{hsieh2023sugarcrepe} and SugarCrepe++~\cite{dumpala2024sugarcrepe++} address this by generating fluent and plausible negatives via LLMs, yet these samples are not explicitly controlled at the level of individual compositional elements, making it difficult to diagnose which element a model actually struggles with.

\noindent \textbf{Hard Negatives for Compositional Reasoning.}
Several methods improve compositional reasoning by augmenting contrastive training with hard negatives.
NegCLIP~\cite{yuksekgonul2022and} constructs hard negatives by perturbing word order and attribute bindings.
StructureCLIP~\cite{huang2024structure} generates semantic negatives via structure-guided word swapping and extends the model architecture to incorporate an additional scene graph input.
TripletCLIP~\cite{patel2024tripletclip} adopts LLM-generated contradictory captions and synthesized negative images~\cite{sauer2024adversarial}, but does not explicitly control which compositional element is modified.
As a result, these methods provide uneven supervision across compositional elements, leading to element-specific biases that hinder balanced compositional reasoning.
% In contrast, CS-CLIP localizes each compositional element via scene graphs and constructs fluent, semantically contradictory hard negatives through a generate-and-rank strategy, providing balanced supervision across all elements.

% \section{Controlled Analysis of Compositional Reasoning}
\section{Controlled Analysis}
\label{sec:controlled_analysis}
% To diagnose how compositionality-aware VLMs handle different compositional elements, we construct controlled test cases that isolate one element at a time. We then evaluate how each model discriminates between positive and hard negative captions across objects, attributes, and relations.

\subsection{Construction of Test Cases}
\label{sec:subsec:construction_testcases}

We construct test cases from VG-Relation and VG-Attribute~\cite{yuksekgonul2022and}, targeting three compositional elements: objects, attributes, and relations. For each dataset, we extract global vocabularies of objects, relations, and attributes from the scene graph annotations, defining the candidate spaces for controlled perturbations.

For each positive caption, we generate a hard negative by modifying exactly one compositional element at a time. For VG-Relation, positives follow the template ``\{subject\} is \{relation\} \{object\}''; we replace either the relation or the object by sampling from the corresponding vocabulary, yielding the Relation and Object subsets. For VG-Attribute, positives follow ``the \{attribute$_1$\} \{object$_1$\} and the \{attribute$_2$\} \{object$_2$\}''; we substitute one attribute by sampling from the attribute vocabulary, yielding the Attribute subset. We use random sampling to avoid biasing the evaluation toward any specific hard negative generation method. We manually filter synonyms and near-duplicates to reduce ambiguous negatives. Illustrative examples are shown at the bottom of Figure~\ref{fig:core_analysis}, and further construction details are provided in Appendix~\ref{sec:appendix:testcases}.

\subsection{Evaluating Discriminative Confidence}
We consider two qualities: \emph{accuracy} and \emph{sensitivity}. Accuracy is the proportion of cases where a model assigns a higher score to the positive caption than to its negative counterpart. Sensitivity captures how confidently the model separates positives from negatives, measured by Cohen's $d$~\cite{cohen2013statistical}: 
\begin{equation}
    d = (\mu_{\text{pos}} - \mu_{\text{neg}}) / s_{\text{pooled}},
\end{equation}
where $\mu_{\text{pos}}$ and $\mu_{\text{neg}}$ are the mean similarity scores for positive and negative captions and $s_{\text{pooled}}$ is the pooled standard deviation. Robust VLMs should achieve both high accuracy and high sensitivity (\ie, high Cohen’s $d$) consistently across all compositional elements, indicating confident discrimination on each element rather than gains concentrated on a few.

\begin{figure*}[th!]
    \centering
    % 너비: 텍스트너비, 높이: 7cm
    % \framebox[\textwidth]{\rule{0pt}{7cm}}
    \includegraphics[width=\textwidth]{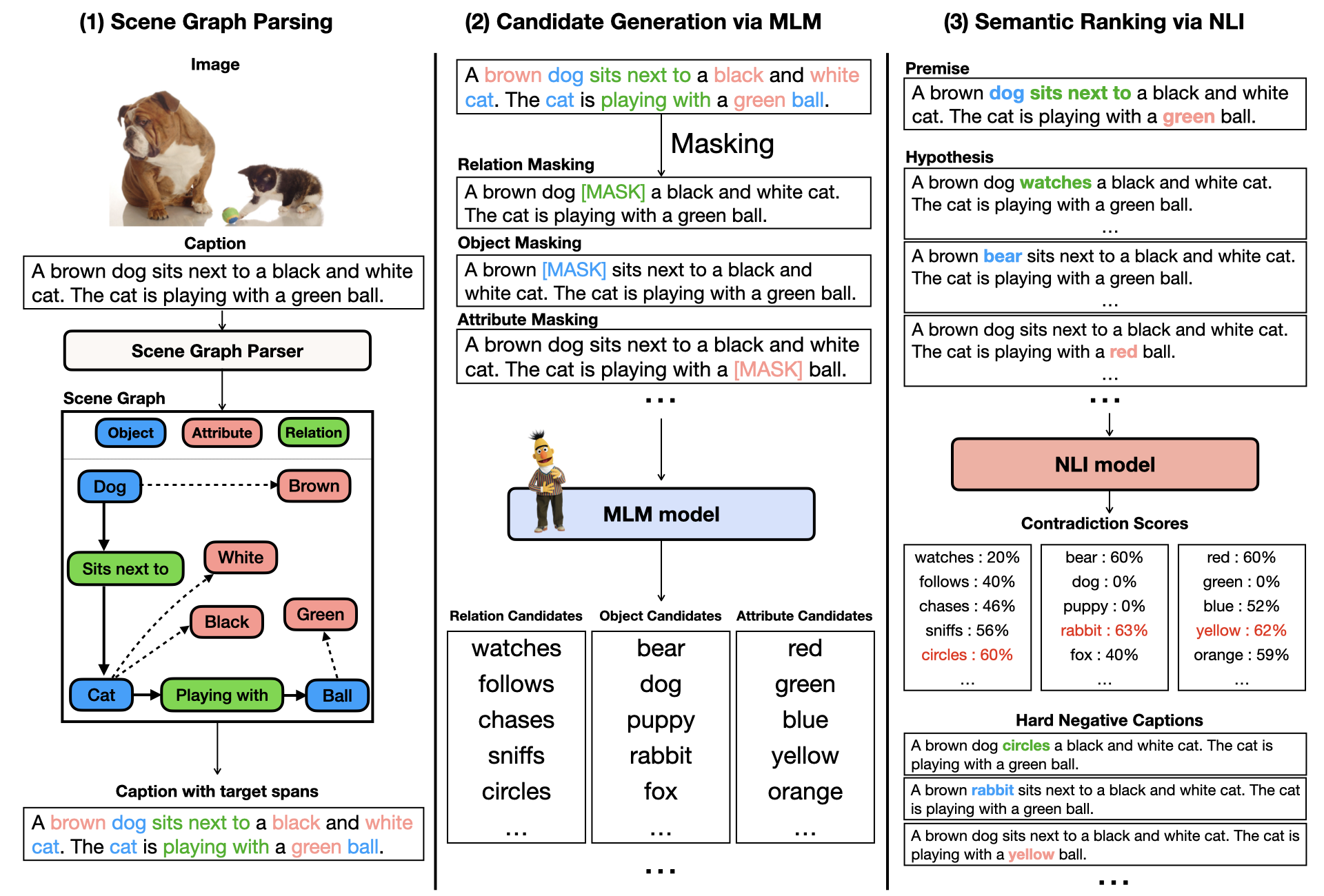}
    \caption{\textbf{Overview of CS-CLIP}. Our framework generates hard negatives in three steps: (1) identifying compositional elements via scene graph parsing, (2) generating candidates with an MLM, and (3) selecting contradictory captions using an NLI model.}
    \label{fig:method}
\end{figure*}

\subsection{Results on our Test Cases}
\label{sec:subsec:analysis_results}
In Figure~\ref{fig:core_analysis} (top), we evaluate vanilla CLIP~\cite{radford2021learning} and three compositionality-aware VLMs: NegCLIP~\cite{yuksekgonul2022and}, StructureCLIP~\cite{huang2024structure}, and TripletCLIP~\cite{patel2024tripletclip}, and report their performance across compositional element changes.

While most VLMs demonstrate improved accuracy over CLIP, this does not necessarily indicate robust compositional reasoning. For instance, StructureCLIP outperforms CLIP for object changes with substantially low sensitivity. More importantly, VLMs fail to demonstrate consistent improvements across element changes. Specifically, they marginally exceed CLIP for object changes; however, they often underperform it for attribute and relation changes. Notably, NegCLIP significantly underperforms CLIP (63.6 $\rightarrow$ 54.4) with substantially lower sensitivity.

% We focus on baselines here; results of CS-CLIP are reported in Section~\ref{sec:experiments}.
% Figure~\ref{fig:core_analysis} (top) shows the results across the three compositional elements.
% On the Object subset, all models marginally exceed CLIP in accuracy, but StructureCLIP shows notably lower sensitivity, indicating that its higher accuracy does not stem from confident discrimination.
% On the Attribute subset, NegCLIP and StructureCLIP achieve larger gains with improved sensitivity, while TripletCLIP fails to improve over CLIP.
% Most strikingly, on the Relation subset, every compositionality-aware VLM underperforms vanilla CLIP in both metrics.

As previously discussed, these deficiencies have not been exposed in existing benchmarks. This suggests that current compositionality-aware VLMs may rely on superficial cues or shortcut correlations rather than genuine compositional reasoning.
% These results indicate strong element-specific biases in existing compositionality-aware VLMs: their gains are concentrated in objects, while relations remain a consistent weakness where they fall behind even vanilla CLIP. 
% We attribute this pattern to a coverage gap in prior hard negative methods, where relation-level perturbations are structurally under-represented compared to objects and attributes, leaving relations under-supervised.
% These observations suggest that addressing element-specific biases requires hard negatives that explicitly target each compositional element with balanced coverage.
% \section{CS-CLIP: Compositional Scene graph-guided CLIP}
\section{Method}
\label{sec:method}

To address the element-specific biases identified in our analysis, we propose Compositional Scene Graph-guided CLIP (CS-CLIP).
CS-CLIP constructs hard negatives that (i) target every compositional element with element-level precision, and (ii) differ from the positive only in the targeted element, without spurious differences such as grammatical artifacts or semantic equivalence.
As illustrated in Figure~\ref{fig:method}, each component of CS-CLIP is chosen to satisfy one of these requirements: scene graph parsing for element-level targeting (i), and a generate-and-rank pipeline combining MLM for fluency and NLI for semantic contradiction (ii).

\subsection{Hard Negative Augmentation}
\label{sec:Hard Negative Augmentation}

\noindent \textbf{Scene Graph Parsing.}
To enable element-level control, we parse each caption $T$ into a scene graph $G = (O, R, A)$ using an off-the-shelf parser~\cite{wu2019unified}, where $O$, $R$, $A$ denote the sets of objects, relations, and attributes. 
The graph localizes each compositional element to its span in $T$, which serves as the target for the generate-and-rank process below.

\noindent \textbf{Candidate Generation via MLM.}
To prevent grammatical artifacts that allow models to bypass compositional reasoning, we use a pre-trained MLM~\cite{devlin2019bert} to generate fluent candidate tokens for the target span. Given a target span in $T$, we mask the span to form $T_{\text{masked}}$ and obtain the top-$K$ candidate tokens ($K=16$):
\begin{equation}
    \mathcal{C}_{cand} = \text{Top-}K_{w \in \mathcal{V}} \left( P_{MLM}(w \mid T_{masked}) \right),
\end{equation}
where $\mathcal{V}$ is the filtered MLM vocabulary. We filter $\mathcal{V}$ to remove sub-word artifacts (e.g., ``\#\#ing'') and to retain only words matching the grammatical category (noun, verb, adjective) of the target span, using WordNet~\cite{miller1995wordnet} for validation.

\noindent \textbf{Semantic Ranking via NLI.}
MLM candidates may include words that preserve the original meaning (e.g., synonyms), yielding semantically equivalent paraphrases rather than true contradictions. To filter these out, we use a pre-trained NLI model~\cite{liu2019roberta} to select the candidate that semantically contradicts the original caption. Treating $T$ as the premise and each candidate $T'_c \in \mathcal{C}_{\text{cand}}$ as the hypothesis, the NLI model assigns a contradiction score $S_{\text{contra}}(T, T'_c)$, and the final hard negative is selected as:
\begin{equation}
T' = \operatorname*{argmax}_{T'_c \in \mathcal{C}_{\text{cand}}} S_{\text{contra}}(T, T'_c).
\end{equation}

\noindent \textbf{Augmentation Types.}
We instantiate the above pipeline as several augmentation types, each perturbing a compositional element in a distinct way:
\begin{itemize}
    \item \textbf{Replace (Object, Attribute, Relation):} We mask and replace the span of an object, attribute, or relation identified by the scene graph, using the full generate-and-rank pipeline.
    \item \textbf{Add (Object, Attribute):} We insert ``and [MASK]'' next to an existing object or attribute and fill the mask via the generate-and-rank pipeline, introducing an additional element to the caption.
    \item \textbf{Swap (Object, Attribute):} We swap two spans of the same element type within the caption via a rule-based procedure. Unlike Replace and Add, Swap does not require MLM or NLI since it operates by rearranging existing words.
\end{itemize}
For each positive caption, we generate one hard negative per applicable type, skipping types whose target elements are absent from the scene graph (e.g., no Add-Attribute negative when the caption has no attribute span).

\subsection{Contrastive Learning Objectives}
\label{sec:training}

We adopt CLIP~\cite{radford2021learning} as our backbone, consisting of a visual encoder and a textual encoder that project images and captions into a shared embedding space. We integrate our hard negatives into training as follows.

\noindent \textbf{Training instances.}
Let $(v, c)$ denote a positive image-caption pair and $c^-$ the embedding of one of its hard negatives. Since each caption can have multiple hard negatives (one per applicable type), we treat each $(v, c, c^-)$ triplet as an independent training instance within the batch.

\begin{table*}[t]
\centering
\resizebox{0.92\textwidth}{!}{
\begin{tabular}{l ccc cc cc c c}
\toprule
\multirow{2}{*}{\textbf{Method}} 
& \multicolumn{3}{c}{\bf Replace} 
& \multicolumn{2}{c}{\bf Swap} 
& \multicolumn{2}{c}{\bf Add} 
& \multirow{2}{*}{\textbf{Avg} $\uparrow$} 
& \multirow{2}{*}{\textbf{Std} $\downarrow$} \\
\cmidrule(lr){2-4} \cmidrule(lr){5-6} \cmidrule(lr){7-8}
& \textbf{Rel.} & \textbf{Obj.} & \textbf{Att.} 
& \textbf{Obj.} & \textbf{Att.} 
& \textbf{Obj.} & \textbf{Att.} 
& & \\
\midrule
\multicolumn{10}{l}{\textit{\textcolor{gray}{Standard VLMs}}} \\
CLIP & 68.9 & 90.9 & 80.0 & 61.3 & 63.6 & 76.8 & 68.3 & 76.3 & 9.6 \\
BLIP-2 & 64.2 & 94.0 & 74.4 & 53.6 & 51.0 & 85.8 & 74.5 & 77.2 & 14.7 \\
\hline
\multicolumn{10}{l}{\textit{\textcolor{gray}{Compositionality-aware VLMs}}} \\
NegCLIP & 76.4 & 92.6 & 85.9 & \underline{75.0} & 75.2 & \underline{88.7} & 82.8 & 84.8 & \underline{6.5} \\
StructureCLIP & 73.8 & 93.5 & 85.6 & 70.3 & \underline{80.4} & 85.4 & 82.6 & 83.8 & 7.2 \\
TripletCLIP & \underline{82.7} & \textbf{94.4} & \underline{86.5} & 67.4 & 72.6 & 87.3 & \underline{85.6} & \underline{85.8} & 8.6 \\
\midrule
\textbf{CS-CLIP (Ours)} 
& \textbf{84.1}\scriptsize{$\pm$0.5} & \underline{94.1}\scriptsize{$\pm$0.4} & \textbf{88.4}\scriptsize{$\pm$0.4} 
& \textbf{76.7}\scriptsize{$\pm$1.2} & \textbf{81.0}\scriptsize{$\pm$0.9} 
& \textbf{91.0}\scriptsize{$\pm$1.2} & \textbf{89.4}\scriptsize{$\pm$0.3} 
& \textbf{88.6}\scriptsize{$\pm$0.2} & \textbf{5.6}\scriptsize{$\pm$0.4} \\
\bottomrule
\end{tabular}}
\caption{\textbf{Results on SugarCrepe.} We report accuracy (\%) for each subtype and the average score (Avg) across all samples from the seven subsets. We additionally report the standard deviation (Std) of the seven subset accuracies as a diagnostic for uniformity across reasoning types; a lower Std indicates smaller type-wise deviation when conditioned on comparable Avg. For CS-CLIP, we report the mean and standard deviation over three random seeds; baselines are evaluated from their public checkpoints.}
\label{tab:sugarcrepe_results}
\end{table*}

\begin{table}[t]
\centering
    \resizebox{\linewidth}{!}{
    \begin{tabular}{l cc cc}
    \toprule
    \multirow{2}{*}{\textbf{Method}} 
    & \multicolumn{2}{c}{\textbf{VG}} 
    & \multicolumn{2}{c}{\textbf{COCO}} \\
    \cmidrule(lr){2-3} \cmidrule(lr){4-5}
    & \textbf{Rel.} & \textbf{Att.} & \textbf{I2T@1} & \textbf{T2I@1} \\
    \hline
    \multicolumn{5}{l}{\textit{\textcolor{gray}{Standard VLMs}}} \\
    CLIP & 58.8 & 60.2 & 50.0 & 30.4 \\
    BLIP-2 & 42.5 & 69.0 & 41.5 & 30.0 \\
    \hline
    \multicolumn{5}{l}{\textit{\textcolor{gray}{Compositionality-aware VLMs}}} \\
    NegCLIP & 78.9 & 70.8 & \textbf{56.2} & \underline{41.5} \\
    StructureCLIP & \underline{81.8} & \textbf{77.6} & \underline{55.5} & 41.4 \\
    TripletCLIP & 51.6 & 60.1 & 39.7 & 33.3 \\
    \midrule
    \textbf{CS-CLIP (Ours)} & \textbf{85.9} & \underline{77.3} & 53.2 & \textbf{42.1} \\
    \bottomrule
    \end{tabular}
    }
    \caption{\textbf{Results on VG-Relation, VG-Attribute, and COCO.} We report accuracy (\%) for VG and Recall@1 (\%) for retrieval tasks in COCO. CS-CLIP achieves the highest accuracy on VG-Relation while remaining competitive on VG-Attribute, and preserves cross-modal retrieval capability on COCO.}
    \label{tab:vg_coco_results}
\end{table}
\begin{table}[t]
    \centering
    \resizebox{0.6\linewidth}{!}{
    \begin{tabular}{l c}
    \toprule
    \textbf{Method} & \textbf{Accuracy} \\
    \midrule
    CLIP & 58.1 \\
    TripletCLIP & 51.9 \\
    NegCLIP & 65.9 \\
    StructureCLIP & \underline{69.1} \\
    \midrule
    \textbf{CS-CLIP (Ours)} & \textbf{71.3} \\
    \bottomrule
    \end{tabular}}
    \caption{\textbf{Zero-shot image-to-text retrieval accuracy on GQA validation split.} CS-CLIP achieves the highest accuracy among all baselines.}
    \label{tab:gqa}
\end{table}
\noindent \textbf{Loss functions.}
We use the standard InfoNCE loss~\cite{oord2018representation} over a batch of $B$ instances, where the contrast is taken across positive image-caption pairs (i.e., other captions in the batch serve as in-batch negatives):

\begin{equation}
\small
\mathcal{L}_{\text{InfoNCE}} =
-\frac{1}{B}\sum_{i=1}^{B} 
\log \frac{\exp(\text{sim}(v_i,c_i))}
{\sum_{j=1}^{B}\exp(\text{sim}(v_i,c_j))},
\end{equation}

\noindent where $\text{sim}(\cdot, \cdot)$ is cosine similarity.

To leverage the hard negatives, we additionally apply a hinge loss~\cite{huang2024structure} on each triplet:

\begin{equation}
\small
\mathcal{L}_{\text{hinge}} = \frac{1}{B} \sum_{i=1}^{B} \max\big(0, \, \omega - \text{sim}(v_i, c_i) + \text{sim}(v_i, c_i^{-})\big),
\end{equation} 

\noindent where $\omega$ is the margin hyper-parameter, which we set to $0.1$. The hinge loss enforces that the positive caption is more similar to the image than its hard negative by at least $\omega$, providing element-targeted supervision that the InfoNCE term alone does not provide.

The final loss combines both:
\begin{equation}
\mathcal{L} = \mathcal{L}_{\text{InfoNCE}} + \mathcal{L}_{\text{hinge}}.
\end{equation}

\section{Experiments}
\label{sec:experiments}

\subsection{Experimental Setup}

\noindent \textbf{Baselines.} We consider two different types of VLMs, including standard VLMs: CLIP~\cite{radford2021learning} and BLIP-2~\cite{li2023blip}, and compositionality-aware VLMs: NegCLIP~\cite{yuksekgonul2022and}, StructureCLIP~\cite{huang2024structure}, and TripletCLIP~\cite{patel2024tripletclip}. We use their publicly available checkpoints\footnote{Since StructureCLIP does not release a public checkpoint, we reproduce it using the official implementation.} and provide details for each baseline in Appendix~\ref{sec:appendix:baseline}.
% for all baselines, except StructureCLIP, which we reproduce using the official implementation. Further baseline details are provided in Appendix~\ref{sec:appendix:baseline}.

\noindent \textbf{Benchmarks.}
We evaluate compositional reasoning on three benchmarks: SugarCrepe~\cite{hsieh2023sugarcrepe}, VG-Relation, and VG-Attribute~\cite{yuksekgonul2022and}. SugarCrepe covers seven reasoning types through LLM-generated hard negatives, providing a broad assessment of compositional reasoning. VG-Relation and VG-Attribute specifically target relation and attribute understanding through hard negatives that swap individual compositional elements. Further benchmark details are provided in Appendix~\ref{sec:appendix:benchmarks}.

\noindent \textbf{Evaluation metrics.}
We report accuracy for all three compositional reasoning benchmarks, measured as the proportion of cases where a model assigns a higher similarity score to the positive caption than to its hard negative.
We additionally evaluate on the COCO 5K split~\cite{lin2014microsoft, karpathy2015deep} for image-to-text retrieval (I2T) and text-to-image retrieval (T2I), reporting Recall@1.

\noindent \textbf{Implementation details.}
We use CLIP (ViT-B/32) as our backbone for fair comparison with the baselines. Hard negatives are generated from the COCO training set~\cite{lin2014microsoft}, yielding approximately 2M training instances. 
Unlike TripletCLIP, which uses additional pre-training data from CC12M~\cite{changpinyo2021conceptual}, we train only on COCO. Negative generation takes approximately 35 hours on a single NVIDIA RTX A6000 48GB GPU. 
We train for 2 epochs with a batch size of 128, using AdamW with an initial learning rate of $10^{-5}$ and cosine decay schedule. 
Training takes approximately 3 hours on the same hardware.

\begin{table*}[th!]
\centering
\resizebox{\textwidth}{!}{
\begin{tabular}{l ccc cc cc c c}
\toprule
\multirow{2}{*}{\textbf{Variant}} 
& \multicolumn{3}{c}{\bf Replace} 
& \multicolumn{2}{c}{\bf Swap} 
& \multicolumn{2}{c}{\bf Add} 
& \multirow{2}{*}{\textbf{Avg}$\uparrow$} 
& \multirow{2}{*}{\textbf{Std}$\downarrow$} \\
\cmidrule(lr){2-4} \cmidrule(lr){5-6} \cmidrule(lr){7-8}
& \textbf{Rel} & \textbf{Obj} & \textbf{Att} 
& \textbf{Obj} & \textbf{Att} 
& \textbf{Obj} & \textbf{Att} 
& & \\
\midrule
\multicolumn{10}{l}{\textit{\textcolor{gray}{Pipeline ablation}}} \\
CLIP 
& 68.9 & 90.9 & 80.0 
& 61.3 & 63.6 
& 76.8 & 68.3 
& 76.3 & 9.6 \\
SG + Random 
& 76.1 & 94.0 & 84.8
& 72.0 & 73.1 
& 92.0 & 80.2 
& 85.3 & 8.2 \\
SG + MLM 
& 76.3 & 93.2 & 85.2 
& 77.2 & 80.1 
& 90.5 & 80.6 
& 85.6 & 6.1 \\
\textbf{SG + MLM + NLI} 
& \textbf{83.7} & \textbf{94.3} & \textbf{87.9} 
& \textbf{78.0} & \textbf{81.9}
& 90.8 & \textbf{89.0} 
& \textbf{88.5} & \textbf{5.2} \\
\midrule
\multicolumn{10}{l}{\textit{\textcolor{gray}{Component leave-one-out}}} \\
\;\; w/o Object negatives 
& 83.8 & \underline{92.4} & 89.7 
& \underline{73.2} & 80.6 
& \underline{90.3} & 88.3 
& 87.9 & 6.3 \\
\;\; w/o Attribute negatives 
& 83.4 & 94.1 & \underline{84.1} 
& 79.7 & \underline{73.6} 
& 91.9 & \underline{84.0} 
& 87.2 & 6.4 \\
\;\; w/o Relation negatives 
& \underline{77.7} & 94.3 & 88.2 
& 75.2 & 82.4 
& 90.4 & 89.9 
& 87.4 & 6.6 \\
\bottomrule
\end{tabular}}
\caption{\textbf{Ablation study on SugarCrepe.}
We report pipeline and component-level ablations along with element-wise standard deviation (Std).
Underlined values indicate the subtypes directly affected by removing the corresponding component-specific negatives.}
\label{tab:ablation}
\end{table*}

\subsection{Main Results}
% \noindent \textbf{Element-wise Compositional Reasoning.}
\noindent \textbf{Robust compositional reasoning across element changes.}
We examine whether CS-CLIP addresses the element-specific biases identified in Section~\ref{sec:controlled_analysis}.
As shown in Figure~\ref{fig:core_analysis} (top), CS-CLIP achieves the highest accuracy and sensitivity across all three compositional elements, while baselines underperform vanilla CLIP on relations.
Table~\ref{tab:vg_coco_results} shows the same pattern on VG, where CS-CLIP reaches state-of-the-art on VG-Relation and remains competitive on VG-Attribute.
These results suggest that CS-CLIP provides balanced supervision across compositional elements, mitigating the element-specific biases in existing VLMs. \\

\noindent \textbf{Gains across reasoning and element types.}
Table~\ref{tab:sugarcrepe_results} shows that CS-CLIP achieves the highest average accuracy and the lowest type-wise standard deviation (Std) on SugarCrepe.
A lower Std indicates more uniform performance across the seven reasoning types.
Notably, TripletCLIP achieves a strong average yet shows a deviation nearly unchanged from vanilla CLIP, indicating that its gains do not translate into robust reasoning across types.
NegCLIP and StructureCLIP show similarly uneven patterns.
CS-CLIP, in contrast, performs best or second-best on every reasoning type, suggesting that its average gain reflects robust compositional reasoning rather than improvement on certain types.

\begin{table}[t!]
\centering
\resizebox{0.9\linewidth}{!}{
\begin{tabular}{l c c}
\toprule
\textbf{Variant} & \textbf{Placement} & \textbf{Avg} \\
\midrule
CS-CLIP-noSG & Random position & 85.9 \\
\textbf{CS-CLIP (Ours)} & SG-guided target span & \textbf{88.5} \\
\bottomrule
\end{tabular}}
\caption{\textbf{Effect of structural placement on SugarCrepe.} Both variants share the same vocabulary produced by our pipeline and differ only in placement.}
\label{tab:vocab_control}
\end{table}

\noindent \textbf{Generalization beyond Compositional Benchmarks.}
We evaluate whether CS-CLIP's compositional improvements preserve general representation capabilities and transfer to downstream tasks.
On COCO cross-modal retrieval (Table~\ref{tab:vg_coco_results}), CS-CLIP achieves competitive performance on both text-to-image and image-to-text retrieval, while TripletCLIP degrades substantially despite its compositional gains.
To further assess downstream transferability, we construct a zero-shot retrieval task on the GQA validation split~\cite{hudson2019gqa}.
For each of 10,000 image-question-answer triples, we construct a positive text by concatenating the question and the correct answer, and a negative text by replacing the correct answer with another object or attribute from the image's scene graph annotation.
The model selects the text with higher similarity to the image embedding (Appendix~\ref{app:gqa}).
As shown in Table~\ref{tab:gqa}, CS-CLIP achieves the highest accuracy among all baselines.
These results indicate that CS-CLIP preserves cross-modal alignment and transfers to downstream visual reasoning.

\subsection{Analysis}
\label{sec:analysis}

\noindent \textbf{Effect of Each Pipeline Stage.}
Table~\ref{tab:ablation} (top) reports a pipeline ablation.
Starting from CLIP, we sequentially add scene graph-guided target spans, MLM-based candidate generation, and NLI-based ranking, with average accuracy rising from 76.3 to 88.5 and type-wise std dropping from 9.6 to 5.2.
Scene graph guidance drives the largest accuracy gain, while MLM and NLI further refine the negatives, each contributing additional reductions in type-wise deviation.

\noindent \textbf{Component-Level Contributions.}
Table~\ref{tab:ablation} (bottom) reports a leave-one-out ablation where we remove the hard negatives corresponding to one compositional element at a time.
Removing any element type degrades performance, but the drop is concentrated on the subtypes directly involving that element (underlined). 
For example, removing Relation negatives causes the largest drop on Replace-Relation, while removing Attribute negatives most affects Replace- and Swap-Attribute. 
This pattern indicates that element-targeted supervision directly determines element-wise performance, supporting our design choice of generating hard negatives for every compositional element.
\begin{figure*}[t]
    \centering
    \includegraphics[width=\linewidth]{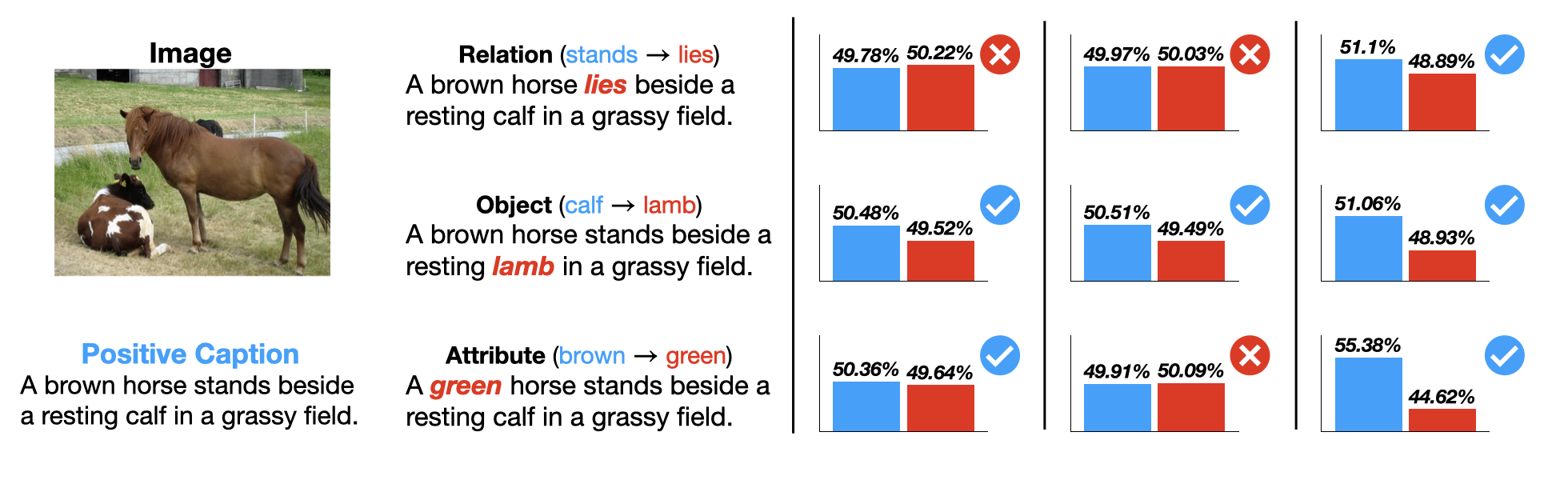}
    \caption{\textbf{Qualitative analysis.} We present an image with its corresponding positive and negative captions, along with prediction probabilities across models. $\checkmark$ (\textcolor{blue}{blue}) and $\times$ (\textcolor{red}{red}) denote correct and incorrect answers, respectively. Compared to CLIP and TripletCLIP, CS-CLIP demonstrates a clearer margin in distinguishing compositional element changes and selects the correct answers.
    }
    \label{fig:qualitative}
\end{figure*}

\begin{figure}[t]
    \centering
    % \framebox[\linewidth]{\rule{0pt}{4cm}}
    \includegraphics[width=\linewidth]{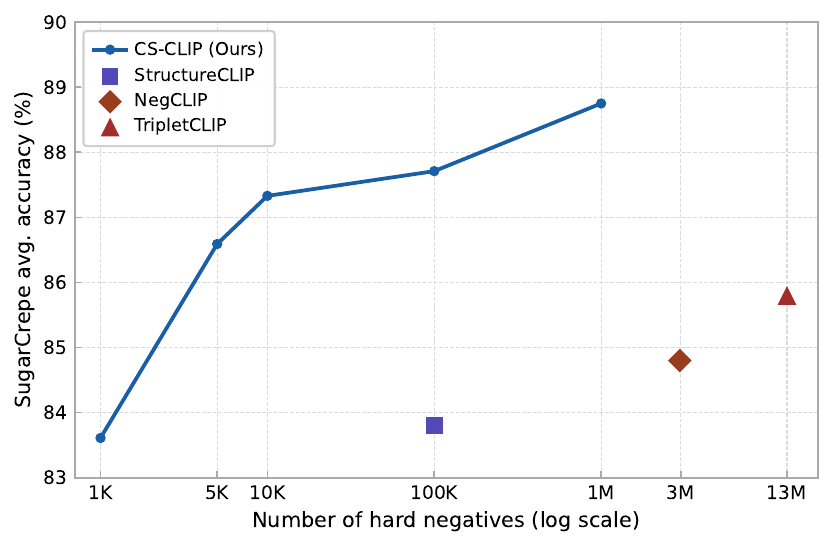}
    \caption{\textbf{Sample efficiency of CS-CLIP on SugarCrepe.} CS-CLIP is trained across data scales from 1K to 1M hard negatives. Baseline checkpoints are placed at their reported training scales.}
    \label{fig:sample_efficiency}
\end{figure}

\noindent \textbf{Vocabulary Diversity vs.\ Structural Placement.}
To test whether CS-CLIP's gains come from the diverse vocabulary produced by our pipeline or from where the perturbation is placed, we compare two variants with the same vocabulary but different placements (Table~\ref{tab:vocab_control}).
CS-CLIP-noSG places perturbations at random positions, while CS-CLIP places them at scene graph-guided spans.
The structural placement yields a substantial gain, suggesting that the benefit comes primarily from precise element-level control rather than vocabulary alone.

\noindent \textbf{Sample Efficiency.}
We train CS-CLIP across data scales from 1K to 1M hard negatives sampled from our generated pool, providing a controlled within-method scaling experiment (Figure~\ref{fig:sample_efficiency}).
Baseline checkpoints are plotted at their publicly reported training scales; since these methods differ in training data and implementation, the baseline points serve as descriptive context rather than a controlled cross-method comparison.
The within-method curve shows that CS-CLIP maintains its performance with substantially fewer hard negatives, reaching strong accuracy well before the largest scale.
This suggests that the gains of CS-CLIP stem from the quality and targeting of its hard negatives rather than data scale.

% \noindent \textbf{Downstream Transferability.}
% We test whether the representation learned by CS-CLIP transfers beyond compositional benchmarks by evaluating zero-shot image-to-text retrieval on the GQA validation split~\cite{hudson2019gqa} (Table~\ref{tab:gqa}).
% For each of 10,000 randomly sampled image-question-answer triples, the model chooses between the correct text (question + correct answer) and a negative text in which the correct answer is replaced with another entity from the image's scene graph annotation.
% Further details on the evaluation setup are provided in Appendix~\ref{app:gqa}.
% CS-CLIP achieves the highest accuracy among all baselines, indicating that element-targeted training improves the underlying representation in a way that generalizes to downstream visual reasoning without task-specific fine-tuning.

\subsection{Qualitative Analysis}
Figure~\ref{fig:qualitative} illustrates how models handle individual compositional element changes. For Object changes, all models distinguish positive and negative captions. 
On Relation cases (e.g., ``stands'' replaced with ``lies''), CLIP and TripletCLIP assign higher scores to the negative, while CS-CLIP correctly identifies the positive. 
On Attribute cases (e.g., ``brown horse'' replaced with ``green horse''), baselines are misled by background colors such as a green grass field, suggesting reliance on global color cues rather than the targeted object. 
CS-CLIP remains confident in the correct attribute, indicating that its training successfully grounds attributes to their corresponding objects.

\section{Conclusion}
We revealed that existing compositionality-aware VLMs suffer from element-specific biases, indicating a lack of robust compositional reasoning.
To address this, we proposed CS-CLIP, which constructs element-targeted hard negatives through scene graph parsing combined with masked language modeling and natural language inference.
By explicitly targeting each compositional element and ensuring that hard negatives are fluent yet semantically contradictory to the positive, CS-CLIP forces the model to rely on compositional reasoning rather than superficial cues or shortcut correlations.
Across compositional reasoning benchmarks, CS-CLIP achieves robust performance across reasoning types, preserves cross-modal capability, and transfers to downstream visual reasoning, with strong sample efficiency.
We will release our code and trained models to facilitate future research.

% Our findings suggest that explicitly controlling which compositional element is perturbed is essential for compositional reasoning in vision-language models.

% We identified that compositionality-aware VLMs achieve uneven gains across reasoning types due to element-specific biases, revealing a lack of balanced compositional reasoning.  
% To address this, we proposed CS-CLIP, an element-targeted hard negative augmentation framework built on scene graph parsing, MLM, and NLI. 
% Experiments demonstrate that CS-CLIP achieves balanced compositional reasoning across all elements, preserves general cross-modal capability, and transfers to downstream visual reasoning. 
% Our findings suggest that explicitly controlling which compositional element is perturbed is essential for compositional reasoning in vision-language models.

\section{Limitations}
While our study demonstrates the effectiveness of CS-CLIP, we acknowledge several limitations. Our framework relies on an off-the-shelf scene graph parser, MLM and NLI modules, meaning any inherent linguistic biases in these modules may propagate to the generated negatives. However, this modularity allows CS-CLIP to scale with future advancements in these modules, improving performance without requiring architectural changes. Our negative generation relies solely on original captions. Since these captions describe the visual content, the derived negatives remain visually relevant. While incorporating visual features could further refine this process, our current approach serves as an efficient and effective alternative.

% This study has several limitations. First, our evaluation and training are restricted to the English language, as the benchmarks and pre-trained models used (e.g., CLIP, BERT) are primarily English-centric. Second, the quality of the generated hard negatives relies on the performance of off-the-shelf scene graph parsers and NLI models; any errors in these components could introduce noise during training. Third, while we focus on three core compositional elements (objects, relations, and attributes), more complex compositional aspects such as temporal reasoning and negation were not addressed. Finally, the proposed method was evaluated on a specific set of benchmarks, and its performance on broader, open-domain visual reasoning tasks remains to be further explored.

% dependency of off-the-shelf modules, not utilizaing image for hard negative generation, under explored utilization of compostional reasoning enhanced multimodal representation, 
% Furthermore, due to limited computational resources, we were unable to evaluate our method on larger backbones such as CLIP (ViT-L/14), leaving the verification of its scalability to larger models for future work. 
% Finally, while we demonstrate significant gains on standard compositional reasoning benchmarks, the utility of our enhanced representations in broader downstream tasks like vision-language navigation remains under-explored and warrants further investigation.

\section{Ethical Considerations}
Our research utilizes publicly available datasets, including Visual Genome and COCO, and follows their respective terms of use. The proposed CS-CLIP framework aims to improve the compositional reasoning capabilities of vision-language models. However, like other models derived from CLIP, CS-CLIP may inherit societal, racial, or gender biases present in large-scale web-crawled training data. While our approach focuses on improving compositional accuracy, it does not explicitly filter or mitigate these underlying biases. Therefore, caution should be exercised when deploying this model in sensitive real-world applications. No personal or private data was collected or used during this study.

\section*{Acknowledgments}
 This work was supported by the Hyundai Motor Chung Mong-Koo Foundation. This work was partly supported by grants from the IITP (RS-2021-II211343-GSAI/10\%, RS-2022-II220951-LBA/10\%, RS-2022-II220953-PICA/10\%, RS-2026-25553157-MIACC/10\%), the NRF (RS-2024-00353991-SPARC/15\%, RS-2023-00274280-HEI/15\%), the KEIT (RS-2025-25453780/15\%), and the KIAT (RS-2025-25460896/15\%), funded by the Korean government. 


\bibliography{custom}

\newpage

\appendix

\section{Baseline Details}
\label{sec:appendix:baseline}
% CLIP, BLIP2, NegCLIP, StructureCLIP, TripletCLIP
We compare CS-CLIP against two representative vision-language models (VLMs), CLIP and BLIP-2, as well as three state-of-the-art compositionality-aware VLMs: NegCLIP, StructureCLIP, and TripletCLIP. We report results using publicly released checkpoints and the official code provided by the authors.

\textbf{CLIP}~\cite{radford2021learning}: A contrastive vision-language pre-training method that aligns image and text representations by predicting cross-modal correspondences. We use the official OpenAI CLIP (ViT-B/32) checkpoint. 

\textbf{BLIP-2}~\cite{li2023blip}: An efficient strategy that bootstraps from frozen image encoders and large language models via a Querying Transformer (Q-Former). BLIP-2 was pre-trained on the same large-scale datasets as its predecessor, BLIP~\cite{li2022blip}, totaling approximately 129 million image-text pairs. We use the publicly released BLIP-2 checkpoint.

\textbf{NegCLIP}~\cite{yuksekgonul2022and}: A CLIP variant that utilizes composition-aware hard negative mining (e.g., perturbing word order or attribute bindings) during training to reduce ``bag-of-words'' behavior. NegCLIP is trained on the COCO training dataset~\cite{lin2014microsoft}.

\textbf{StructureCLIP}~\cite{huang2024structure}: A framework that integrates explicit scene graph knowledge (objects, attributes, relations) to generate structured negatives. It employs a knowledge-enhanced encoder to refine multimodal representations. We reproduced this model using the official implementation.

\textbf{TripletCLIP}~\cite{patel2024tripletclip}: A strategy that augments data with synthetic hard negative captions (via in-context learning) and images (via text-to-image models) and uses an alternating triplet-style contrastive objective. It is pre-trained on CC12M~\cite{changpinyo2021conceptual} and fine-tuned on the COCO training dataset.

\section{Benchmark Details}
\label{sec:appendix:benchmarks}

\textbf{SugarCrepe}~\cite{hsieh2023sugarcrepe}.
SugarCrepe is a vision-language compositionality benchmark formulated as a binary image-to-text retrieval test. For each image, the model must choose the positive caption that correctly describes the image over a minimally perturbed hard negative caption. The hard negative differs only by small compositional changes.
SugarCrepe is constructed from COCO image-text pairs and generates hard negatives via an LLM-based rewriting procedure, followed by human validation to remove false negatives and an adversarial refinement step to reduce text-only artifacts.
The benchmark covers multiple fine-grained hard-negative families, including Replace (object/attribute/relation), Swap (object/attribute), and Add (object/attribute).

\textbf{VG-Relation}~\cite{yuksekgonul2022and}.
VG-Relation is part of the Attribution--Relation--Order (ARO) benchmark built from Visual Genome annotations~\cite{krishna2017visual}.
Each test instance contains a cropped image region tightly covering two salient objects and a pair of candidate captions of the form ``$X$ \emph{relation} $Y$'' versus the swapped-order caption ``$Y$ \emph{relation} $X$'' (e.g., ``the horse is eating the grass'' vs.\ ``the grass is eating the horse'').
The benchmark spans a diverse set of relations (including spatial prepositions and verb relations).

\textbf{VG-Attribute}~\cite{yuksekgonul2022and}.
VG-Attribute evaluates whether a model correctly binds object properties to the right entities.
Each test instance is a cropped region containing two objects and two candidate captions that swap the attributes between the objects (e.g., ``the crouched cat and the open door'' vs.\ ``the open cat and the crouched door'').

\textbf{COCO Retrieval}~\cite{lin2014microsoft}.
For standard MSCOCO image--text retrieval, we follow the widely adopted Karpathy split~\cite{karpathy2015deep}: 113,287 images for training, 5,000 for validation, and 5,000 for testing, with each image paired with five reference captions.
We evaluate both text-to-image retrieval (T2I) and image-to-text retrieval (I2T) on the Karpathy test split.
We report Recall@1, defined as the proportion of queries for which at least one ground-truth match appears in the top-1 retrieved results.

\section{Test Cases Details}
\label{sec:appendix:testcases}
Our controlled test cases are constructed from the densely structured scene graph annotations of the Visual Genome (VG) dataset, which provide object, relation, and attribute information for each image. Specifically, we utilize the VG-Relation and VG-Attribute subsets, and generate negative captions by modifying only a single compositional element of the original positive captions (i.e. object, relation, or attribute). To support systematic replacement, we build global vocabularies for each compositional element. The object vocabulary is collected from scene graph object annotations and contains 899 unique objects, while the attribute vocabulary is compiled separately and includes 541 unique attributes. In addition, following the same procedure, we construct a relation vocabulary consisting of 91 unique relations. We manually filter synonyms and near-duplicates from each vocabulary to ensure that sampled substitutes produce semantically distinct negatives.

\textbf{Relation}: In the VG-Relation dataset~\cite{yuksekgonul2022and}, positive captions follow the template “{subject} is {relation} {object}.” We replace the target relation with a randomly sampled substitute from the relation vocabulary. This yields 13,595 pairs in which discrimination depends solely on the relation (e.g., changing “man is riding horse” to “man is standing next to horse”).

\textbf{Object}: Similarly derived from the VG-Relation dataset using the same template, we replace the target object with a substitute sampled from the object vocabulary. This yields 23,937 pairs in which discrimination depends solely on the object (e.g., changing “man is riding horse” to “man is riding car”).

\textbf{Attribute}: In the VG-Attribute dataset~\cite{yuksekgonul2022and}, positive captions follow the template ``the \{attribute$_1$\} \{object$_1$\} and the \{attribute$_2$\} \{object$_2$\}.'' We replace the target attribute with a randomly sampled substitute from the attribute vocabulary while keeping the remaining context unchanged. This process yields 28,748 pairs in which discrimination depends solely on the attribute (e.g., changing ``the brown giraffe and the green tree'' to ``the blue giraffe and the green tree'').

\section{Details of the GQA Downstream Evaluation}
\label{app:gqa}

We build a zero-shot image-to-text retrieval benchmark on the GQA balanced validation split~\cite{hudson2019gqa} to test whether a model grounds a scene-consistent answer over a fluent but visually-incorrect distractor.

\subsection{Constructing the Two-Choice Set}
Each instance is a triplet $\langle \text{image},\, t^{+},\, t^{-}\rangle$ in which the model must select the positive text $t^{+}$ over the negative text $t^{-}$.
Both texts are prefixed with the question so the two choices differ only in the answer content:
\begin{align}
t^{+} &= [\,q \;\Vert\; f\,], \\
t^{-} &= [\,q \;\Vert\; \tilde{f}\,],
\end{align}
where $q$ is the question, $f$ is the gold \texttt{fullAnswer}, and $\tilde{f}$ replaces the gold answer span in $f$ with a different entity drawn from the same image's scene graph.
The replacement is sampled from object names and attributes present in the scene graph, excluding the objects referenced by the question and answer annotations.
This keeps $t^{-}$ linguistically plausible yet visually false, so the model must rely on fine-grained grounding rather than language priors.

\subsection{Quality Filtering}
We discard yes/no answers, since a binary flip is trivially separable.
We reject a replacement whose character-level similarity to the gold answer exceeds $0.85$, or that is a substring of the gold answer with a length difference of at most two characters, removing near-duplicates such as singular/plural variants.
We reject a negative sentence whose similarity to the positive sentence exceeds $0.9$, ensuring the two choices describe different content.
After filtering, we randomly sample $10{,}000$ instances with a fixed seed for reproducibility.

\subsection{Evaluation Protocol}
For each instance we encode the image and both texts, $\ell_2$-normalize the embeddings, and compute cosine similarities $s^{+}=\langle z_{\text{img}}, z_{t^{+}}\rangle$ and $s^{-}=\langle z_{\text{img}}, z_{t^{-}}\rangle$.
The instance is correct if $s^{+} > s^{-}$, and we report accuracy:
\begin{equation}
\text{Acc} = \frac{1}{N}\sum_{i=1}^{N} \mathbb{1}\!\left[\, s^{+}_{i} > s^{-}_{i} \,\right], \quad N = 10{,}000,
\end{equation}
\noindent where chance is $50\%$.
No GQA data is used for training.

\begin{table}[t]
\centering
\small
\begin{tabular}{p{\dimexpr0.5\linewidth-2\tabcolsep\relax} p{\dimexpr0.5\linewidth-2\tabcolsep\relax}}
\toprule
\textbf{Positive text $t^{+}$} & \textbf{Negative text $t^{-}$} \\
\midrule
\textit{The boy in the bed is looking at what animal? The boy is looking at the cat.} &
\textit{The boy in the bed is looking at what animal? The boy is looking at the sky.} \\
\addlinespace
\textit{What vegetable is in the pasta that is in the top? The vegetable is broccoli.} &
\textit{What vegetable is in the pasta that is in the top? The vegetable is peas.} \\
\bottomrule
\end{tabular}
\caption{Examples of the constructed two-choice instances.}
\label{tab:gqa_examples}
\end{table}

\begin{figure*}[th!]
    \centering
    % \framebox[\textwidth]{\rule{0pt}{9cm}}
    \includegraphics[width=\linewidth]{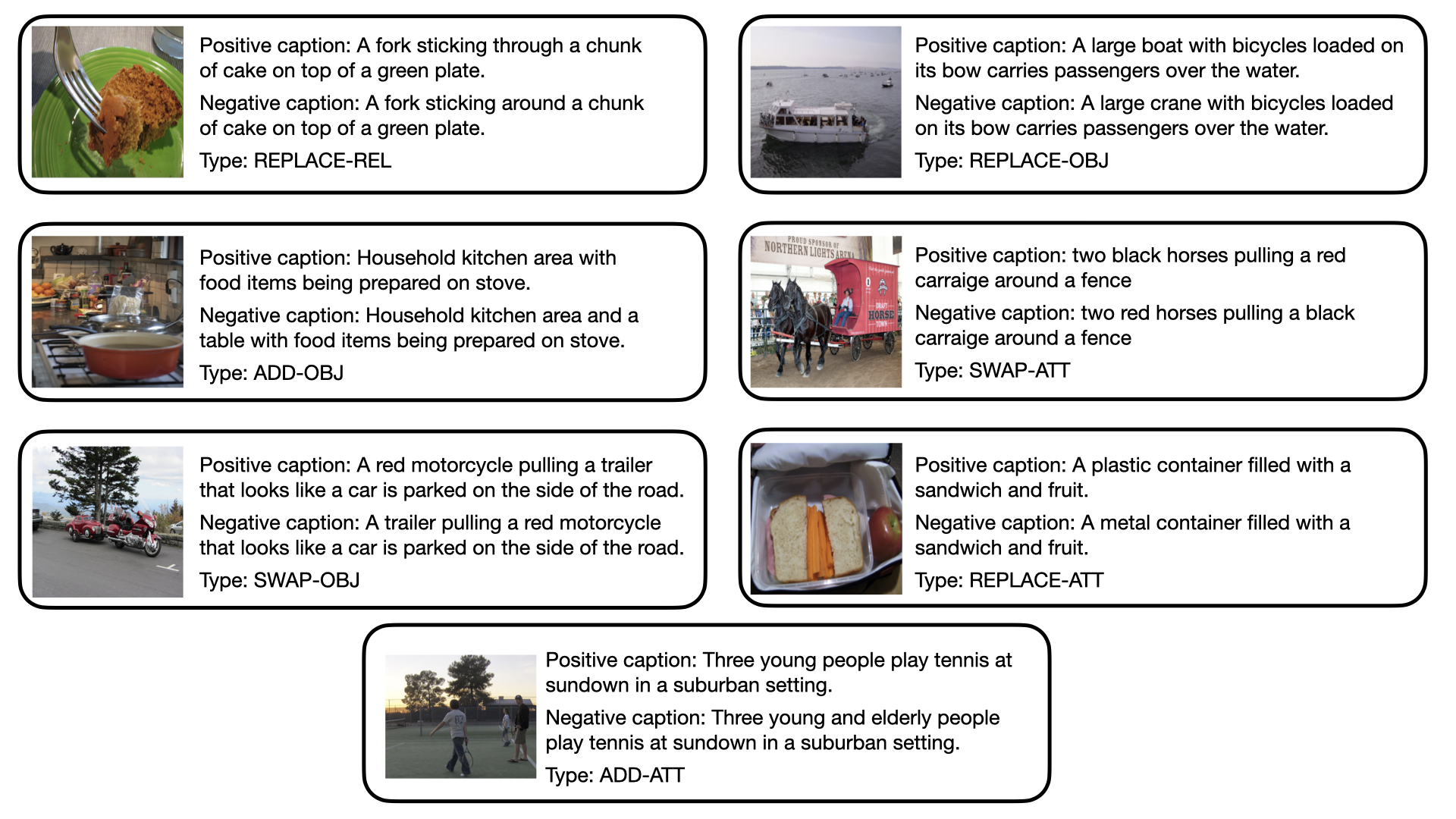}
    \caption{\textbf{Examples of CS-CLIP's Hard Negatives.} }
    \label{fig:hn_examples}
\end{figure*}

\section{Effectiveness on Different Backbones}
\label{sec:appendix:backbones}

To evaluate the generalizability of our approach, we additionally apply our hard negative augmentation to StructureCLIP~\cite{huang2024structure} as a backbone.
StructureCLIP incorporates a graph encoder that takes scene graphs as input.
We adopt its architecture but train it using our hard negatives.

\begin{table}[h!]
\centering
\resizebox{\linewidth}{!}{
\begin{tabular}{l l ccc cc cc c}
\toprule
\multirow{2}{*}{\textbf{Backbone}} 
& \multirow{2}{*}{\textbf{Ours}}
& \multicolumn{3}{c}{\bf Replace} 
& \multicolumn{2}{c}{\bf Swap} 
& \multicolumn{2}{c}{\bf Add} 
& \multirow{2}{*}{\textbf{Avg} $\uparrow$} \\
\cmidrule(lr){3-5} \cmidrule(lr){6-7} \cmidrule(lr){8-9}
& & \textbf{Rel} & \textbf{Obj} & \textbf{Att} 
& \textbf{Obj} & \textbf{Att} 
& \textbf{Obj} & \textbf{Att} & \\
\midrule
\multirow{2}{*}{CLIP}
& -- & 68.9 & 90.9 & 80.0 & 61.3 & 63.6 & 76.8 & 68.3 & 76.3 \\
& \checkmark & 83.7 & 94.3 & 87.9 & 78.0 & 81.9 & 90.8 & 89.0 & \textbf{88.5} \\
\midrule
\multirow{2}{*}{StructureCLIP}
& -- & 73.8 & 93.5 & 85.6 & 70.3 & 80.4 & 85.4 & 82.6 & 83.8 \\
& \checkmark & 83.3 & 94.0 & 88.3 & 78.8 & 80.9 & 90.3 & 86.2 & \textbf{88.0} \\
\bottomrule
\end{tabular}}
\caption{\textbf{Effectiveness across different backbones.}
We apply our hard negative augmentation to both CLIP and StructureCLIP backbones.
Our augmentation consistently improves performance on both backbones.}
\label{tab:appendix_backbone}
\end{table}

As shown in Table~\ref{tab:appendix_backbone}, applying our hard negatives improves the average accuracy of CLIP from 76.3\% to 88.5\% and StructureCLIP from 83.8\% to 88.0\%.
This indicates that our augmentation provides consistent supervision across different backbone architectures, and the benefits of CS-CLIP are not tied to a specific encoder design.

% \section{Additional Experiments}
% \label{sec:appendix:extraexps}
% % 정성평가 그림에 다른모델 결과도 포함

\section{Examples of CS-CLIP's Hard Negatives}
\label{sec:appendix:extraexps}
% 생성한 하드네거티브 예시들
Figure~\ref{fig:hn_examples} shows representative examples of hard negatives generated by CS-CLIP across all seven augmentation types.
Each negative caption differs from the positive only in the targeted compositional element (e.g., Replace-Relation changes ``through'' to ``around''), while preserving fluency and the rest of the sentence structure.

\end{document}